\documentclass[preprint,3p]{elsarticle}

\usepackage{amssymb}
\usepackage{amsmath}

\usepackage{algorithm}
\usepackage{algpseudocode}
\usepackage{bm}
\usepackage{makecell}
\usepackage{tikz}
\usetikzlibrary{shapes.geometric}
\usetikzlibrary{calc, shapes, backgrounds, fit, positioning}
\usetikzlibrary{decorations.pathmorphing}

\definecolor{C0}{HTML}{E41A1C}
\definecolor{C1}{HTML}{377EB8}
\definecolor{C2}{HTML}{984EA3}
\definecolor{C3}{HTML}{FF7F00}
\definecolor{C4}{HTML}{CCCC33}
\definecolor{C5}{HTML}{F781BF}
\definecolor{C6}{HTML}{4DAF4A}
\definecolor{C7}{HTML}{A65628}

\newcommand{\fail}[1]{p_{#1}(\textrm{failure})}

\usepackage{lineno}

\journal{Information Fusion}

\begin{document}

\begin{frontmatter}



\title{Interpretable Rules for Online Failure Prediction: A Case Study on the Metro do Porto dataset}


\author[label1]{Matthias Jakobs\corref{cor1}}
\ead{matthias.jakobs@tu-dortmund.de}
\cortext[cor1]{Corresponding author}
\author[label2,label3]{Bruno Veloso}
\ead{bveloso@fep.up.pt}
\author[label2,label3]{Jo\~ao Gama} 
\ead{jgama@fep.up.pt}

\affiliation[label1]{organization={Lamarr Institute for Machine Learning and Artificial Intelligence, TU Dortmund University},
            city={Dortmund},
            country={Germany}}
\affiliation[label2]{organization={INESC-TEC},
            city={Porto},
            country={Portugal}}

\affiliation[label3]{organization={Faculty of Economics - University of Porto},
            city={Porto},
            country={Portugal}}

\begin{abstract}
Due to their high predictive performance, predictive maintenance applications have increasingly been approached with Deep Learning techniques in recent years. However, as in other real-world application scenarios, the need for explainability is often stated but not sufficiently addressed. This study will focus on predicting failures on Metro trains in Porto, Portugal. While recent works have found high-performing deep neural network architectures that feature a parallel explainability pipeline, the generated explanations are fairly complicated and need help explaining why the failures are happening. This work proposes a simple online rule-based explainability approach with interpretable features that leads to straightforward, interpretable rules. We showcase our approach on MetroPT2 and find that three specific sensors on the \textit{Metro do Porto} trains suffice to predict the failures present in the dataset with simple rules.
\end{abstract}




\end{frontmatter}



\section{Motivation}
\label{motivation}

Maintenance approaches in industrial settings today are often sorted into three different categories.
The most straightforward approach, \textit{corrective maintenance}, merely replaces machine parts whenever they break.
Depending on the industry and machine, this will result in unexpected outages or even harm to people.
An alternative is to install a \textit{preventive maintenance} protocol where, for example periodically, old parts are exchanged for new ones, even if they are not broken.
The downsides of this approach are a high cost and potentially little benefit if the old part is not broken, or, even worse, exchanging a working part with a new part with a production error, introducing failures unnecessarily.
In recent years, a third way has become more attractive to practitioners due to the adoption of Machine Learning (ML) methods in Industry 4.0, namely \textit{predictive maintenance}.
Here, the goal is to predict whether or not a failure is about to occur and which part is responsible for the upcoming failure based on historical and real-time data provided by the industry process. 
This approach has the upside of a low cost, given that the prediction method is accurate and has a small likelihood of unnecessarily introducing faulty parts.
Since these predictive maintenance approaches should be usable in the real world and might impact people financially and physically, there is a high need for explainability to understand why a model predicts a failure to be imminent.

This work will focus on predictive maintenance in the context of the trains run by the \textit{Metro do Porto} in Porto, Portugal. 
Specifically, we are concerned with predicting whether or not the central air processing unit (APU) will experience problems since the trains are not equipped with a backup unit, and the unit is crucial for the train to be able to operate.
Recent work has approached this task with Deep Learning methods, specifically Autoencoders (AE), that learn characteristics of non-failure data and can alert the practitioners if newly acquired data significantly deviates from these characteristics.
While these works address the need for explainability using online rule-learning approaches, we find that the extracted rules are often too complicated and do not capture the true characteristics of the underlying data in an interpretable manner.
Thus, our contributions are the following:

\begin{itemize}
    \item We present an online rule-based learning approach that uses the output of a simple Autoencoder architecture to extract interpretable rules that give a clear insight into why both failures in the MetroPT2 dataset occur. Specifically, we find that only one sensor placed on the APU must predict both failures in advance.
    \item Our approach can generate both local rules (explaining each failure individually) and global rules (where both failures are explained jointly). 
    \item These rules are extracted on transformations of the raw sensor values over a long time window, providing a large amount of interpretability in contrast to previous approaches by having interpretable features and concise rules, suggesting that the dataset is less challenging than previously thought. 
\end{itemize}

This paper is structured in the following way: 
First, we discuss the properties of the MetroPT2 dataset used for all experiments in Sec. \ref{sec:metropt}.
Afterwards, we provide an overview of the other approaches for failure prediction on this and similar datasets in Sec. \ref{sec:related}.
Next, we present the methodology of the Autoencoder-based failure prediction approach, the online rule-learning algorithm in Sec. \ref{sec:methodology}, and the experiments and results of both in Sec. \ref{sec:experiments}.
Lastly, we discuss our findings, limitations and future work opportunities (Sec. \ref{sec:discussion}) and give a conclusion and final remarks in Sec. \ref{sec:conclusion}.

\section{Problem definition}
\label{sec:metropt}

This work will focus on the MetroPT2 dataset \cite{MetroPT2} for failure prediction.
The Air Production Unit (APU) installed onto the trains of the \textit{Metro do Porto} is a central unit responsible for various tasks.
For example, the APU controls the opening and closing of the train doors and raising or lowering the train at the station's platform.
This means that failures of the APU will result in the train being unable to continue operation since there is no redundancy built into the train to combat APU failures.

The MetroPT2 dataset contains sensor readings of 16 analogue and digital sensors placed onto the APU between April 28th and July 28th 2022, at a frequency of 1Hz.
The analogue measurements are the following  (descriptions taken from \cite{MetroPT2}):
\begin{itemize}
    \item \texttt{TP2} - Pressure on the compressor (bar)
    \item \texttt{TP3} - Pressure generated at the pneumatic panel (bar)
    \item \texttt{H1} - Valve that is activated when the pressure read by the pressure switch of the command is above the operating pressure of 10.2 bar
    \item \texttt{DV\_pressure} - Pressure exerted due to pressure drop generated when air dryers towers discharge the water. When it equals zero, the compressor works under load (bar).
    \item \texttt{Reservoirs} - Pressure inside the air tanks installed on the trains (bar)
    \item \texttt{Oil\_temperature} - Temperature of the oil present on the compressor ($^\circ C$)
    \item \texttt{Flowmeter} - Airflow was measured on the pneumatic control panel ($m^3/h$)
    \item \texttt{Motor\_Current} - Motor's current, which should present the following values: (i) close to 0A when the compressor turns off; (ii) close to 4A when the compressor is working offloaded; and (iii) close to 7A when the compressor is operating under load (A)
\end{itemize}
In terms of digital sensors, we will only mention the two most important sensors for this work and refer to \cite{MetroPT2} for a complete description:
\begin{itemize}
    \item \texttt{COMP} - Electrical signal of the air intake valve on the compressor. It is active when there is no admission of air on the compressor, meaning that the compressor turns off or is working offloaded.
    \item \texttt{LPS} - A low-pressure signal, activated when the pressure is below 7 bars. This signal is used neither for training nor testing.
\end{itemize}

Even though the dataset is unlabeled, two failures are known for the dataset, given by maintenance reports.
The first failure is an air leak, while the second is an oil leak.
Table \ref{tab:failures} shows these failures, their duration, and the timestamp at which the \texttt{LPS} signal turned on.
The \texttt{LPS} signal is a failure warning already built into the train conductor control panel.
The goal of predicting failures on this dataset is to find the failures two hours before the \texttt{LPS} signal turns on so the train can be safely removed from the tracks in time.

\begin{table*}[]
    \centering
    \begin{tabular}{llll}
         Failure & Start time & End time & \texttt{LPS} Signal \\ \hline
         Air Leak & 2022-06-04 10:19:24 & 2022-06-04 14:22:39 & 2022-06-04 11:26:01 \\
         Oil Leak & 2022-07-11 10:10:18 & 2022-07-14 10:22:08 & 2022-07-13 19:43:52 \\
    \end{tabular}
    \caption{Failures present in MetroPT2 dataset.}
    \label{tab:failures}
\end{table*}

\section{Related Work}
\label{sec:related}
In anomaly detection, the goal is to detect datapoints from a larger set of samples that are significantly different to the other samples.
It is a long-studied problem domain that, especially in recent years, is approached more and more with machine learning and deep learning methods \cite{chandolaAnomalyDetectionSurvey2009,pangDeepLearningAnomaly2021}.
Related to anomaly detection is failure prediction in the field of predictive maintenance, where the prediction task is to determine whether or not a machine, part, etc. is about to have a critical failure (and why).
Specifically, failure prediction is only concerned with critical failures that are non-recoverable, i.e., where the process has to be halted and cannot recover itself.
To tackle this prediction task, as with anomaly detection, machine learning and deep learning is becoming more and more popular \cite{carvalhoSystematicLiteratureReview2019,serradillaDeepLearningModels2022}.
In this work, we want to focus on one particular method to predict failures, namely to observe the reconstruction error of Autoencoders, in the context of detecting failures on the MetroPT2 dataset \cite{MetroPT2}.
In \cite{silva2023predictive} the authors evaluated multiple AE architectures trained on a portion of failure-free data.
The best model, a Wasserstein Autoencoder GAN architecture, achieved a perfect $F_1$ score, meaning no false alarms were signalled, and both failures in the MetroPT2 dataset were found.
In \cite{gamaFaultDetectionAnomaly2024}, the authors had a similar approach but found an LSTM-based Autoencoder to perform best.
In contrast to \cite{silva2023predictive}, who predicted 30-minute windows of data, the authors in \cite{gamaFaultDetectionAnomaly2024} identified compressor cycles using the \texttt{COMP} sensor and predicted each of these cycles as anomalous or not.
In addition, each compressor cycle window was discretized into non-overlapping bins, on which summary statistics such as the mean, min and max values were calculated \cite{davariPredictiveMaintenanceBased2021}.

Since the goal of predictive maintenance and failure prediction is to be applied in real-world settings, the question of why these models learn (and predict) the way they do emerges naturally.
Thus, tools from Explainable Machine Learning are often used to investigate these deep learning models since they are not inherently interpretable \cite{rudinStopExplainingBlack2019}.
One kind of explanations are feature attributions, where the explanation comes in the form of one real value per feature, indicating the \textit{importance} to the prediction, such as LIME \cite{ribeiroWhyShouldTrust2016} and SHAP \cite{lundbergUnifiedApproachInterpreting2017} values.
One downside of these approaches is that it is often unclear and ambiguous how these values should be interpreted.
Additionally, most methods assume feature independence for faster computation, which can lead to problems when investigating time-series data, a data type often found in industry applications.
Thus, previous works on failure prediction on the MetroPT2 dataset favor rule-based explanations \cite{silva2023predictive,gamaFaultDetectionAnomaly2024}.
Rules have the advantage of being very easy to interpret, as long as they are not excessively long and as long as the antecedents (e.g., the features) are interpretable themselves.

The authors of \cite{silva2023predictive,gamaFaultDetectionAnomaly2024} utilized an explainability layer including AMRules \cite{duarteAdaptiveModelRules2016,ribeiroOnlineAnomalyExplanation2023}, an online rule-learning approach to model the Autoencoders reconstruction error based on the raw input values.
Specifically, Chebyshev sampling \cite{aminianChebyshevApproachesImbalanced2021} is used to account for the fact that failures are rare events.
Afterwards, the rules that have the highest support for the failure duration are returned as an explanation for that failure.
We argue that their rule-based approach could be improved by addressing the following two issues:
\begin{itemize}
    \item The rules are trained on individual timesteps from raw data to predict the reconstruction error, i.e., each one-second vector of sensor values is an individual training example. This results in hard-to-interpret rules since the resolution is too fine-grained to find easy, general rules covering the failures.
    \item Choosing the reconstruction error as a target for rule learning results in overcomplicating the rules unnecessarily. We argue that choosing the output of the failure detection pipeline, i.e., $ \fail{} > 0.5 $ as a binary label, will lead to more straightforward to comprehend rules.
\end{itemize}

\section{Methodology}
\label{sec:methodology}

Let $\bm{X} \in \mathbb{R}^{C \times T}$ be a multivariate time series with $C$ channels of length $T$.
We denote with $\bm{X}^{(t)}$ the vector of all $C$ channel values at time $t$, i.e., $\bm{X}^{(t)} \in \mathbb{R}^C$.
We define a \textit{windowing} $\mathcal{X}_{L,d}$ with length $L$ and stride $d$ on $\bm{X}$ via

\begin{align*}
    \mathcal{X}_{L,d} &:= \{ \underbrace{(\bm{X}^{(1)}, \dots, \bm{X}^{(L)})}_{:= \bm{X}_1}, \underbrace{(\bm{X}^{(d)}, \dots, \bm{X}^{(d+L)})}_{:= \bm{X}_2}, \dots \} 
\end{align*}

Thus, $\mathcal{X}_{L,d}$ defines overlapping fixed-length windows of the original time series with a stride (or step size) of $d$ to get a suitable representation for training our Autoencoder model.

\subsection{Failure detection}

Autoencoder (AE) are Neural Network architectures consisting of two parts: The encoder $E: \mathbb{R}^{C \times L} \rightarrow \mathbb{R}^{m \times L}$ learns to project the input data to a $m$-dimensional latent representation.
Afterwards, a decoder $D: \mathbb{R}^{m \times L} \rightarrow \mathbb{R}^{C \times L}$  learns to reconstruct the input from the latent representation.
Thus, we get a reconstruction for a window $\bm{X}_i$ via $\hat{\bm{X}}_i = D(E(\bm{X}_i))$.
The training objective is to minimize the reconstruction error $\ell(\bm{X}_i, \hat{\bm{X}}_i) = \sum (\bm{X}_i-\hat{\bm{X}}_i)^2$ between each training window and its reconstruction $\hat{\bm{X}}_i$.
For anomaly and failure detection, the expected reconstruction error for non-anomalous data is assumed to be significantly smaller than the reconstruction error of failures.
Thus, the reconstruction error can serve as a proxy for the degree of anomaly of a datapoint.

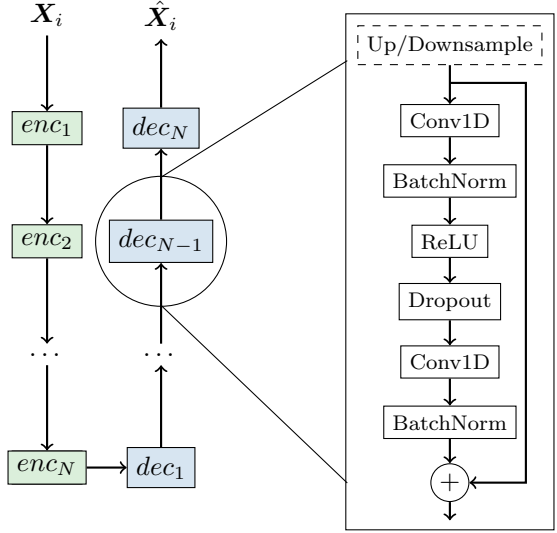
\begin{figure}
    \centering
    \begin{tikzpicture}
        \node[rectangle, node distance = 1.5cm] (x) {$\bm{X}_i$};
        \node[draw, rectangle, fill=C6!20, node distance = 1.5cm, below of=x] (enc1) {$enc_1$};
        \node[draw, rectangle, fill=C6!20, node distance = 1.5cm, below of=enc1] (enc2) {$enc_2$};
        \node[rectangle, node distance = 1.5cm, below of=enc2] (enc3) {$\dots$};
        \node[draw, rectangle, fill=C6!20, node distance = 1.5cm, below of=enc3] (enc4) {$enc_4$};
        \node[draw, rectangle, fill=C6!20, node distance = 1.5cm, below of=enc3] (enc4) {$enc_N$};
        \node[draw, rectangle, fill=C1!20, node distance = 1.5cm, right of=enc4] (dec4) {$dec_1$};
        \node[rectangle, node distance = 1.5cm, above of=dec4] (dec3) {$\dots$};
        \node[draw, rectangle, fill=C1!20, node distance = 1.5cm, above of=dec3] (dec2) {$dec_{N-1}$};
        \node[draw, rectangle, fill=C1!20, node distance = 1.5cm, above of=dec2] (dec1) {$dec_N$};
        \node[rectangle, node distance = 1.5cm, above of=dec1] (reconstruction) {$\hat{\bm{X}}_i$};
    
        \node[draw=black,fit=(dec2),inner sep=0.6ex,circle] (tmp) {};
    
        \node[draw, rectangle, node distance = 3.8cm, right of=dec2] (ReLU) {\footnotesize ReLU};
        \node[draw, rectangle, node distance = 0.8cm, above of=ReLU] (BN) {\footnotesize BatchNorm};
        \node[draw, rectangle, node distance = 0.8cm, above of=BN] (Conv1) {\footnotesize Conv1D};
        \node[draw, rectangle, node distance = 0.8cm, below of=ReLU] (DP) {\footnotesize Dropout};
        \node[draw, rectangle, node distance = 0.8cm, below of=DP] (Conv2) {\footnotesize Conv1D};
        \node[draw, rectangle, node distance = 0.8cm, below of=Conv2] (BN2) {\footnotesize BatchNorm};
        
        \node[rectangle, node distance = 0.5cm, above of=Conv1] (ghost1) {};
        \node[draw, dashed, rectangle, node distance = 0.5cm, above of=ghost1] (UD) {\footnotesize Up/Downsample};
        \node[draw, circle, node distance = 0.8cm, below of=BN2, inner sep=0.05cm] (ghost2) {$+$};
        \node[node distance = 0.35cm, minimum width=0.1cm, minimum height=0.1cm, below of=ghost2] (ghost3) {};
        
        \node[draw=black,fit=(ghost1) (ghost2) (ghost3) (UD) (Conv1) (ReLU) (BN) (DP) (Conv2) (BN2) ,inner sep=1ex,rectangle] (tmp3) {};
        \draw[-> ,thick] (ghost1) +(-0.01, 0) -- +(1, 0) -- +(1, -5.3) -- (ghost2);
        \draw[->, thick] (UD) -- (Conv1);
        \draw[->, thick] (Conv1) -- (BN);
        \draw[->, thick] (BN) -- (ReLU);
        \draw[->, thick] (ReLU) -- (DP);
        \draw[->, thick] (DP) -- (Conv2);
        \draw[->, thick] (Conv2) -- (BN2);
        \draw[->, thick] (BN2) -- (ghost2);
        \draw[->, thick] (ghost2) -- +(0, -0.5);
        \draw[-] (tmp) +(0,0.85) -- ($(tmp3) +(-1.35,2.8)$) ;
        \draw[-] ($(tmp) +(0,-0.85)$) -- ($(tmp3) +(-1.35,-2.8)$) ;
    
        \draw[->, thick] (x) -- (enc1);
        \draw[->, thick] (enc1) -- (enc2);
        \draw[->, thick] (enc2) -- (enc3);
        \draw[->, thick] (enc3) -- (enc4);
        \draw[->, thick] (enc4) -- (dec4);
        \draw[->, thick] (dec4) -- (dec3);
        \draw[->, thick] (dec3) -- (dec2);
        \draw[->, thick] (dec2) -- (dec1);
        \draw[->, thick] (dec1) -- (reconstruction);

    \end{tikzpicture}
    \caption{The autoencoder used in this work consists of $N$ encoder and decoder blocks (left) with identical block structures (right).}
    \label{fig:architecture}
\end{figure}

Next, we present our AE architecture in Fig. \ref{fig:architecture} for failure prediction.
The architecture consists of $N$ encoder and decoder blocks; each comprises convolutional layers, batch normalization layers, ReLU activation functions, and Dropout layers.
Optionally, an up- or downsample layer (realized with a fully connected layer) is used where needed to match shapes.
We also utilize skip-connections proposed by \cite{heDeepResidualLearning2016} to improve training.

By using the boxplot method on the training reconstruction error distribution \cite{ribeiroSequentialAnomaliesStudy2016}, we define an anomaly threshold 
$$\tau_{\textrm{anom}} = \beta q_{99}$$
where $q_{99}$ is the $99$th percentile of the distribution of training reconstruction errors and $\beta$ is a hyperparameter.
If the error for a 30-minute slice of test data exceeds $\tau_{\textrm{anom}}$, the slice is considered anomalous.
To reduce the number of false-positive alarms, a low-pass filter is applied to this binary output according to Eq. \ref{eq:lpf}, where $y_t$ is the observed testing slice prediction (i.e., anomalous or not) at time $t$ and $\alpha$ is a smoothing hyperparameter.
Lastly, a failure is detected if the smoothed output exceeds a threshold of $\tau_{\textrm{fail}} = 0.5$.
This reduces the false-positive rate because only prolonged occurrence of anomalous data points should be considered a failure.
From now on, we will refer to the smoothed output (at time $t$) as the probability of failure $\fail{t}$.

\begin{equation}
    \label{eq:lpf}
    z_t = z_{t-1} + \alpha (y_t - z_{t-1}), ~ z_0 = y_0
\end{equation}

\subsection{Online rule-learning approach}

With our model able to detect failures by monitoring the reconstruction error, we will present our approach for extracting interpretable rules for each detected failure.
The idea is to monitor the probability of failure $\fail{}$ and decide, based on this value, whether or not an individual window is part of a failure.
We propose to compute aggregated values for each window $\bm{X}_i$ to get more interpretable rules.
Let $\varphi: \mathbb{R}^{L \times C} \rightarrow \mathbb{R}^{M \times C}$ be a transformation which aggregates the time dimension of length $L$ using $M$ aggregation functions of type $\phi: \mathbb{R}^{L} \rightarrow \mathbb{R}$.
In this work, we utilize each windows' variance, minimum, maximum and mean value for higher interpretability.
Finally, let $\dot{\bm{X}}_i := \varphi(\bm{X}_i)$ be a transformed window.

Our online rule-learning algorithm is shown in Alg. \ref{alg:rules}.
We will start with an empty ruleset $\bm{r} = \{ \}$.
We assume that there are $T$ overlapping windows of data to process and denote the set of indices with $[T] := \{ 1, 2, \dots, T \}$.
Inspired by the approach presented in \cite{gamaRecurrentConceptsData2014}, we define three different states of operation during inference.
First, if $\fail{t} < \tau_{\textrm{warn}}$, where $\tau_{\textrm{warn}}$ is a hyperparameter, we are in the \texttt{Normal} state, i.e., the reconstruction error at time $t$ is considered small.
In this state, we add the current window $\dot{\bm{X}}_t$ to a history of normal data $H$.
The history will hold onto non-anomalous data as predicted by our Autoencoder model.
Once $\fail{t} > \tau_{\textrm{warn}}$, we change to the \texttt{Warning} state.
During the warning state, we add $\dot{\bm{X}}_t$ to a buffer $B$ of anomalous data instead.
Should we eventually reach the \texttt{Failure} state ($\fail{t} > \tau_{\textrm{fail}}$ with $\tau_{\textrm{fail}} > \tau_{\textrm{warn}}$) we create a labeled dataset from $B$ and $H$.
Each example in $B$ is labelled as \textit{failure}, while all examples in $H$ are labelled as \textit{no failure}.
Using all rules in $\bm{r}$ we test whether they cover entirely the labelled datapoints.
If a rule $r \in \bm{r}$ does not cover all labelled data, it is removed from the rule set.
If the rule set is empty, we train decision trees on the labelled dataset to a perfect fit and save them to the rule set.
We train multiple trees since (especially for a small number of anomalous examples) there might be multiple rules that cover all labelled examples.

\begin{figure}
    \centering
    \includegraphics{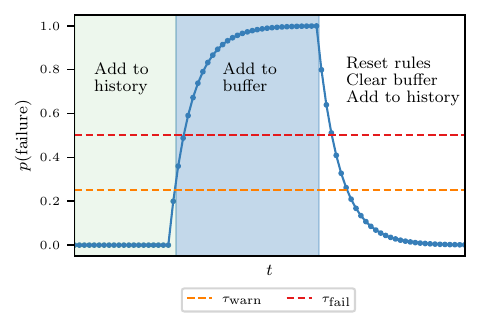}
    \caption{Visualization of our online rule-learning approach. During the green section, we add non-anomalous data to our history. During the blue section, we add the data to the buffer instead and learn fitting rules. Once $\fail{}$ stops monotonically increasing, we reset and return the learned rules, clear the buffer of anomalous data and add the datapoints to the history again. }
    \label{fig:rule-schematic}
\end{figure}

We consider one failure completed if the probability of failure starts decreasing again.
Recall that $\fail{}$ results from low-pass filtering the time series of binary failure / non-failure decisions.
Thus, if $\fail{}$ stops monotonically increasing, the underlying binary decisions are consecutive non-failure predictions, and we assume the failure to have stopped.
See Fig. \ref{fig:rule-schematic} for a schematic overview of this approach.

\begin{algorithm}
\caption{Online rule-learning on failure probability} \label{alg:rules}
    \begin{algorithmic}[1]

        \State $B, H, \bm{r} = \{ \}$ 
        \For{$t\in[T]$}
            \State $\texttt{{\color{black} isIncreasing}} = \fail{t} > \fail{t-1}$
            \State $\texttt{{\color{black} isWarning}} = \fail{t} > \tau_{\textrm{warn}}$
            \State $\texttt{{\color{black} isFailure}} = \fail{t} > \tau_{\textrm{fail}}$
            \If{$\texttt{{\color{black} isWarning}} \wedge \texttt{{\color{black} isIncreasing}}$}
                \State \texttt{{\color{gray} // Enrich buffer}}
                \State $B = B \cup \{ \dot{\bm{X}}_t \}$
            \Else 
                \State \texttt{{\color{gray} // Enrich history instead}}
                \State $B = \{\}$
                \State $H = H \cup \{ \dot{\bm{X}}_t \}$
            \EndIf
            \If{$\texttt{{\color{black} isFailure}}$}
                \If{$\texttt{{\color{black} isIncreasing}}$}
                    \State $x, y = \text{CollectExamples}(B, H)$
                    \State $\bm{r} = \text{UpdateRules}(x, y, \bm{r})$
                \Else
                \State \texttt{{\color{gray} // Return and reset rules}}
                    \State $\bm{r} = \{ \}$
                \EndIf
            \EndIf
        \EndFor
    \end{algorithmic}
\end{algorithm}

\begin{table}[]
    \centering
    \begin{tabular}{ll}
         Failure & Rule \\ \hline
         Air Leak & $\texttt{Flowmeter\_max} > 16.05$ \\
         Oil Leak & $\texttt{Flowmeter\_max} > 16.18$ \\
    \end{tabular}
    \caption{Extracted rules using all available features.}
    \label{tab:rules_flowmeter}
\end{table}

\section{Experiments}
\label{sec:experiments}
Next, we will present the results of our AE architecture for failure detection and the extracted rules for each failure from our proposed online rule-learning approach.
Our code is publicly available on Github \footnote{https://github.com/MatthiasJakobs/metro-xai}.

We formulate the following research questions:
\begin{itemize}
    \item \textbf{Q1}: Can we detect both failures two hours before the \textrm{LPS} signal activates?
    \item \textbf{Q2}: Which sensors of the APU are most useful in detecting the failures?
    \item \textbf{Q3}: Can we defer simple, interpretable rules for both failures?
\end{itemize}

\subsection{Failure detection}
We use the following approach based on \cite{silva2023predictive} for preprocessing the dataset.
The dataset is first split into two parts: A training period (until 2022-06-01) that is considered free of anomalies and failures and a testing period (from 2022-06-01).
Next, the data is transformed into overlapping windows of 30 minutes ($L=1800$) with a stride of 5 minutes ($d=300$) to simulate that we receive new sensor values during a real-world application every $5$ minutes.
The last $30\%$ of training windows are used to measure the validation error during training and estimate $\tau_{\textrm{anom}}$ to prevent overfitting to the training data.
We normalized each sensor channel by predicting mean and standard deviation over the train set and using it to center the remaining data windows.

In terms of Autoencoder architecture, we parameterize the architecture from Sec. \ref{sec:methodology} in the following way:
We utilize $N=10$ temporal blocks for both the encoder and decoder.
All convolution layers have a kernel size of $3$ and are made up of $30$ kernels each.
To increase the receptive field of the encoder and decoder, we increase the dilation of each temporal block exponentially, i.e., each encoder blocks $ I \in [N]$ is assigned a dilation of $2^i$ (and analogous for the decoder).
The embedding dimension was experimentally set to $32$.
Additionally, we set the Dropout probability to $0.2$ for all Dropout layers in the AE to prevent overfitting. 
We train our model using the \texttt{Adam} optimizer \cite{kingmaAdamMethodStochastic2015} with a learning rate of $10^{-4}$ for a total of $200$ epochs.

\begin{figure*}
    \centering
    \includegraphics{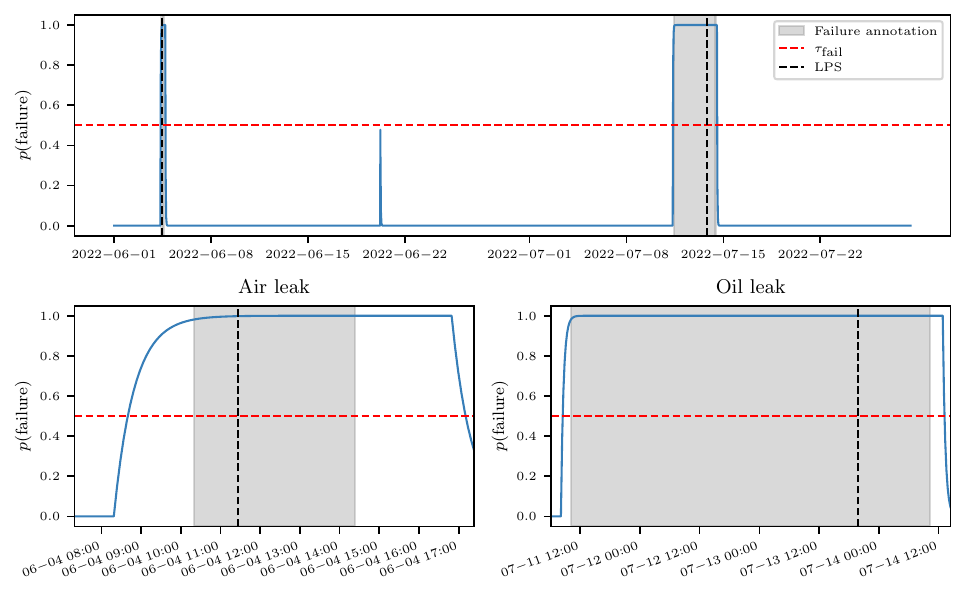}
    \caption{Our models predicted $\fail{}$ (in blue) over the testing period of MetroPT2 (top row) and for both failures in more detail (bottom row). The actual failure periods are shown in grey, as well as the activation of the \texttt{LPS} signal as the dotted black lines. The threshold $\tau_{\textrm{fail}}$ at which $\fail{}$ is reported as a failure is set to $\tau_{\textrm{fail}} = 0.5$ and shown in red.}
    \label{fig:prob_results}
\end{figure*}

After fitting the model to the training data, we found that a smoothing factor $\alpha = 0.15$ and an anomaly threshold of $\tau_{\textrm{anom}} = 3q_{99}$ lead to a perfect $F_1$ score, i.e., both failures are detected (more than) $2$ hours before the respective \texttt{LPS} signal turned on and there were no false positives. 
We show the results of our method in Fig. \ref{fig:prob_results}.
In the top row, our model $\fail{}$ is very small except for both failure cases (and one slightly anomalous section during late June), suggesting that we highly fit the training data without overfitting it.
In the bottom row, we show the two failure periods in more detail.
For the Air leak (left), we note that $\fail{} > 0.5$ starts at 8:43 AM while the \texttt{LPS} signal does not turn on until 11:26 AM, meaning that we detect the failure over $150$ minutes before the \texttt{LPS} signal.
For the Oil leak (right), we detected the failure at 08:32 AM on July 11th, over two full days before the \textrm{LPS} signal turned on. 
This answers research question \textbf{Q1}.

\subsection{Explaining detected failures}
\begin{figure*}
    \centering
    \includegraphics{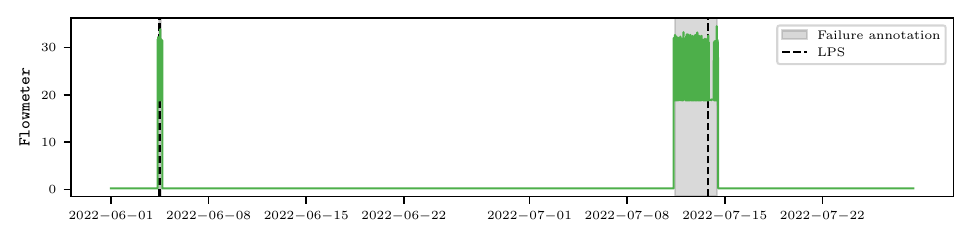}
    \caption{The \texttt{Flowmeter} sensor during the entirety of the test data range. Notice that, most of the time, the values of \texttt{Flowmeter} are minimal and only increase during the annotated failures. For a more detailed view of both failures, see Fig. \ref{fig:flowmeter_combined}. }
    \label{fig:flowmeter_test}
\end{figure*}

\begin{figure}[!t]
    \centering
    \includegraphics{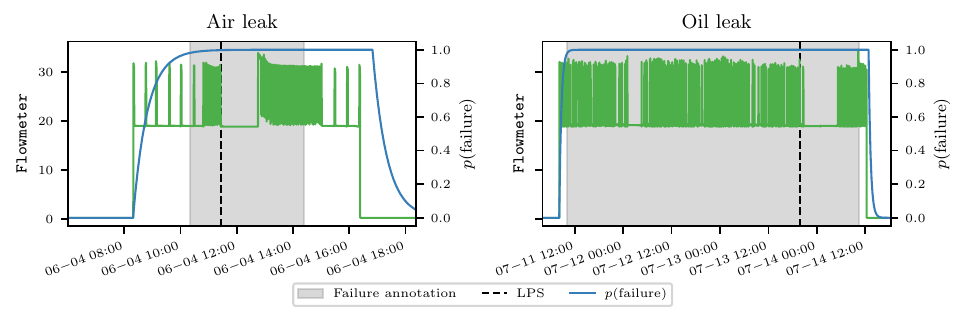}
    \caption{The \texttt{Flowmeter} sensor during the Air and Oil leaks. \texttt{Flowmeter} values are shown in green while the $\fail{}$ is shown in blue. A clear correlation between failures and high values of \texttt{Flowmeter} makes this sensor highly predictive of these failures.}
    \label{fig:flowmeter_combined}
\end{figure}

Next, we apply our online rule-learning algorithm on $\fail{t}$ estimated by our Autoencoder model to investigate probable causes of both detected failures.
We are able to recover one rule for each failure, covering the respective failure period.
The rules are shown in Tab. \ref{tab:rules_flowmeter}.
Interestingly, there seems to be one highly predictive sensor, namely \texttt{Flowmeter}, which is enough to predict whether or not the train will experience a failure.
Specifically, if the value in a $30$ minute slice of data at any point exceeds a threshold of roughly $16 m^3/h$ then this slice is part of a failure.
Notice that this is true for both failures.
Due to our rule-learning algorithm, no false positives are associated with preceding data, i.e., the found rules do not produce false positives for all windows preceding the respective failure.
The strong correlation between high values of \texttt{Flowmeter} and the annotated failures is also apparent when investigating the raw data visually, as can be seen in Fig. \ref{fig:flowmeter_test} and Fig. \ref{fig:flowmeter_combined}.
We can observe that \texttt{Flowmeter} usually has small values, with sporadic peaks aligning perfectly with the observed failures.
We also want to stress that multiple valid thresholds could be used in this case since the difference between failure and no failure value ranges is very large.
Thus, as long as this sensor is present in the APU, early failure detection appears to be a relatively simple process that does not require complex deep-learning methods.
Indeed, as can be observed in Fig. \ref{fig:flowmeter_combined}, the probability of failure (shown in blue) increases at the same time as the raw values of \texttt{Flowmeter}, suggesting a high reliance of our model on this sensor (in particular, a high reconstruction error of that sensor channel).

The correlation between high values of the \texttt{Flowmeter} sensor with both failures answer research questions \textbf{Q2} and \textbf{Q3}.
However, we want to note that \texttt{Flowmeter} is a fairly expensive sensor to install, and not all trains of the \textit{Metro do Porto} in the real world can be equipped with this sensor.
Thus, we also intended to find suitable rules not dependent upon \texttt{Flowmeter}.
After removing the \texttt{Flowmeter} sensor feature from the data used for rule-learning, we were able to find rules that focus on \texttt{Oil\_temperature}, \texttt{Motor\_current} and \texttt{TP2} (see Tab. \ref{tab:rules_noflowmeter}).
These sensors are significantly cheaper to produce and install and thus might be more suitable to be rolled out into every train of \textit{Metro do Porto}.

\begin{table}[]
    \centering
    \begin{tabular}{ll}
         Failure & Rule \\ \hline
         Air Leak & $\texttt{Oil\_temperature\_max} > 72.16$ \\
         Oil Leak & \begin{tabular}[t]{@{}l@{}} $\texttt{TP2\_max} > 10.61 ~ \vee$ \\ $(\texttt{TP2\_max} \leq 10.61 ~ \wedge$ \\ $\texttt{Motor\_current\_min} \leq 0.01)$ \end{tabular} \\
    \end{tabular}
    \caption{Extracted rules using all available features except for \texttt{Flowmeter}.}
    \label{tab:rules_noflowmeter}
\end{table}

To further investigate the rules shown in Tab. \ref{tab:rules_noflowmeter}, we investigated the purity of each Decision Tree node that generated these rules.
We found for the Oil leak that the \texttt{Motor\_current} feature is only necessary to separate one window from the other $53,486$ windows, meaning that the simple rule $\texttt{TP2\_max} > 10.61$ could be considered a suitable explanation for the Oil leak.
The ability to also detect both failures using the \texttt{Motor\_current}, \texttt{Oil\_temperature} and \texttt{TP2} sensors gives more insights into research question \textbf{Q2}.

\subsection{Global rules}
Lastly, we want to investigate whether we can find interpretable global rules that work for both failures to understand better what constitutes anomalous and non-anomalous data.
We slightly change the Algorithm presented in Alg. \ref{alg:rules} by adding the buffer windows after each failure is completed to a separate global buffer.
Thus, at the end of the available test data, we train a Decision Tree to classify all failure windows (found in the global buffer) from all non-failure windows (found in the history).

As seen in Fig. \ref{fig:flowmeter_test}, finding a global rule for both failures is effortless when considering the \texttt{Flowmeter} sensor.
Our algorithm found the following rule that covers all failure windows and did not produce any false positives:
$$
\texttt{Flowmeter\_max} > 9.58 \Rightarrow \textrm{Failure}
$$
When omitting \texttt{Flowmeter} we learned the Decision Tree shown in Fig. \ref{tab:rules_noflowmeter}.
Similar to the local rule found for the Oil leak, we want to stress that the split on \texttt{TP2} is only necessary to separate two windows from the remaining $57,148$ windows.
Thus, we argue that \texttt{Motor\_current} and \texttt{Oil\_temperature} can be considered sufficient to find all failures, answering research question \textbf{Q3} with regards to simple, globaly applicable rules for MetroPT2.

\begin{figure}[!t]
    \centering
    \includegraphics{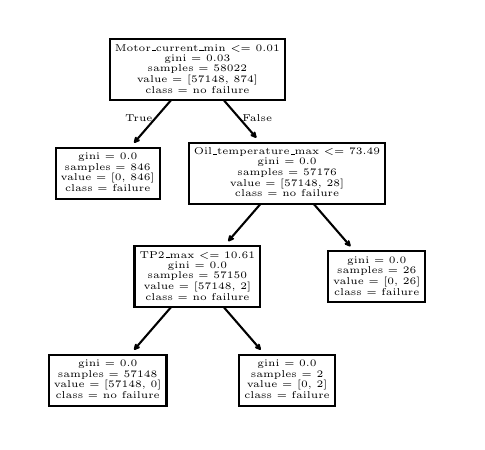}
    \caption{Decision Tree that separates both failures in MetroPT2 from non-failure data without relying on the \texttt{Flowmeter} sensor.}
    \label{fig:noflowmeter_globalrules}
\end{figure}

\section{Discussion}
\label{sec:discussion}

After investigating the extracted rules as well as the raw data itself, we conclude that the MetroPT2 dataset, as presented in \cite{MetroPT2}, is not very challenging for failure and novelty prediction due to the high correlation between values of \texttt{Flowmeter} and the presence of failures in the APU.
Thus, we suggest that future work which targets the MetroPT2 dataset remove this sensor from the data before applying their methods.
Initial experiments from our side suggested that modelling $\fail{}$ using Autoencoders without providing \texttt{Flowmeter} to the model is a substantially more complex problem.
While we could find rules that do not involve \texttt{Flowmeter} that still fit all slices of failure windows, it is tough to learn the notion of thresholding maximum and minimum values with Convolutional Autoencoder architectures.
The dependence on \texttt{Flowmeter} is also why we struggled to fit a good Autoencoder onto MetroPT3 \cite{velosoMetroPTDatasetPredictive2022}, since there is no \texttt{Flowmeter} sensor available.
Initial experiments showed many false-positive failure predictions, suggesting that different methods than our Convolutional Autoencoder to model $\fail{}$ would be necessary.
Thus, we recommend that future work for predictive maintenance on \textit{Metro do Porto} data should also focus on MetroPT3 due to its more challenging nature.

One explicit limitation of our work is that an infinite number of rules could be retrieved from the data.
For instance, choosing as a global threshold for \texttt{Flowmeter} the value $9.59$ instead of $9.58$ (as shown previously) will still result in a perfect $F_1$ score. 
Thus, we cannot confidently gain knowledge about potential breaking points of parts of the APU from these rules.
Another limitation is the correct optimization of hyperparameters such as the smoothing parameter $\alpha$ and boxplot threshold $\beta q_i$. 
While we observe that a range of values for these hyperparameters produces satisfactory results, optimizing with validation data containing actual failures would be a more thorough way of setting these values.
Lastly, in the current iteration, the amount of history saved by our online rule-learning algorithm is unbounded.
While this results in rules that fit all previous non-failure windows perfectly, this solution will not scale in real-world scenarios where months and years of data could be gathered.
We suggest that future work investigate different known strategies to keep a maximum history size, such as random subsampling or removing windows from further in the past with higher probability.
Additionally, since this history is merely there to serve as examples for non-failure data windows, clustering methods could be highly efficient in finding representative windows that summarize large amounts of data from history.

\section{Conclusion}
\label{sec:conclusion}
In this work, we investigated the scenario of failure prediction on a popular real-world dataset from \textit{Metro do Porto}, namely MetroPT2, using Convolutional Autoencoders and an explainability pipeline which extracts interpretable local rules for each failure, as well as interpretable global rules covering all available data.
Using our online rule-learning approach, we found that, in its current state, MetroPT2 is not difficult to solve due to the existence of one highly predictive sensor, namely \texttt{Flowmeter}.
However, we also found rules that do not depend on this highly predictive sensor that are still very easy and interpretable, suggesting that future work should focus on more challenging versions of these datasets, such as MetroPT3.

\subsection*{Acknowledgement}
This research has been funded by the Federal Ministry of Education and Research of Germany and the state of North Rhine-Westphalia as part of the Lamarr Institute for Machine Learning and Artificial Intelligence.









\end{document}